\documentclass[10pt,twocolumn,letterpaper]{article}

\usepackage{cvpr}
\usepackage{times}
\usepackage{epsfig}
\usepackage{graphicx}
\usepackage{amsmath}
\usepackage{amssymb}
\usepackage{xcolor}
\usepackage{mydefs}

\usepackage{url}            
\usepackage{booktabs}       
\usepackage{amsfonts}       
\usepackage{nicefrac}       
\usepackage{microtype}      

\usepackage{times}
\usepackage{epsfig}
\usepackage{graphicx}
\usepackage{amsmath}
\usepackage{amssymb}

\usepackage{units}
\usepackage{multirow}
\usepackage{bm}
\usepackage{mydefs}
\usepackage{bbm}
\usepackage{pifont}
\usepackage{xspace} 

\usepackage{graphicx,calc} 
\usepackage{color}    
\usepackage{ifthen}
\usepackage{paralist}
\usepackage{times}
\usepackage{longtable}
\usepackage{colortbl}
\usepackage{nomencl}
\usepackage{bm}
\usepackage{float}

\usepackage[pagebackref=true,breaklinks=true,letterpaper=true,colorlinks,bookmarks=false]{hyperref}

\cvprfinalcopy 

\def\cvprPaperID{6266} 

\def\eg{\emph{e.g}\onedot} 
\def\ie{\emph{i.e}\onedot}

\def\wrt{w.r.t\onedot}

\newcommand{\supsecref}[1]{\secref{#1}}

\ifcvprfinal\fi
\begin{document}

\title{SurfelGAN: Synthesizing Realistic Sensor Data for Autonomous Driving}

\author{
  \hspace{-1.3cm}
  \begin{tabular}[t]{c}
    Zhenpei Yang$^{1*}$, Yuning Chai$^2$, Dragomir Anguelov$^2$, Yin Zhou$^2$,  Pei Sun$^2$, \\ Dumitru Erhan$^3$, Sean Rafferty$^2$, Henrik Kretzschmar$^2$ \\
    $^1$UT Austin, $^2$Waymo, $^3$Google Brain\\
\end{tabular}
}

\maketitle

\ifcvprfinal
\let\thefootnote\relax\footnotetext{$^*$Work done as an intern at Waymo. Correspondence: yzp@utexas.edu}
\thispagestyle{empty}
\fi
                             
\begin{abstract}
Autonomous driving system development is critically dependent on the ability to replay complex and diverse traffic scenarios in simulation. In such scenarios, the ability to accurately simulate the vehicle sensors such as cameras, lidar or radar is essential. However, current sensor simulators leverage gaming engines such as Unreal or Unity, requiring manual creation of environments, objects and material properties. Such approaches have limited scalability and fail to produce realistic approximations of camera, lidar, and radar data without significant additional work.

In this paper, we present a simple yet effective approach to generate realistic scenario sensor data, based only on a limited amount of lidar and camera data collected by an autonomous vehicle. Our approach uses texture-mapped surfels to efficiently reconstruct the scene from an initial vehicle pass or set of passes, preserving rich information about object 3D geometry and appearance, as well as the scene conditions. We then leverage a SurfelGAN network to reconstruct realistic camera images for novel positions and orientations of the self-driving vehicle and moving objects in the scene. We demonstrate our approach on the Waymo Open Dataset and show that it can synthesize realistic camera data for simulated scenarios. We also create a novel dataset that contains cases in which two self-driving vehicles observe the same scene at the same time. We use this dataset to provide additional evaluation and demonstrate the usefulness of our SurfelGAN model.
\end{abstract}

\section{Introduction}
\label{Section:Introduction}

\begin{figure*}[tp]
  \centering
\includegraphics[width=\textwidth]{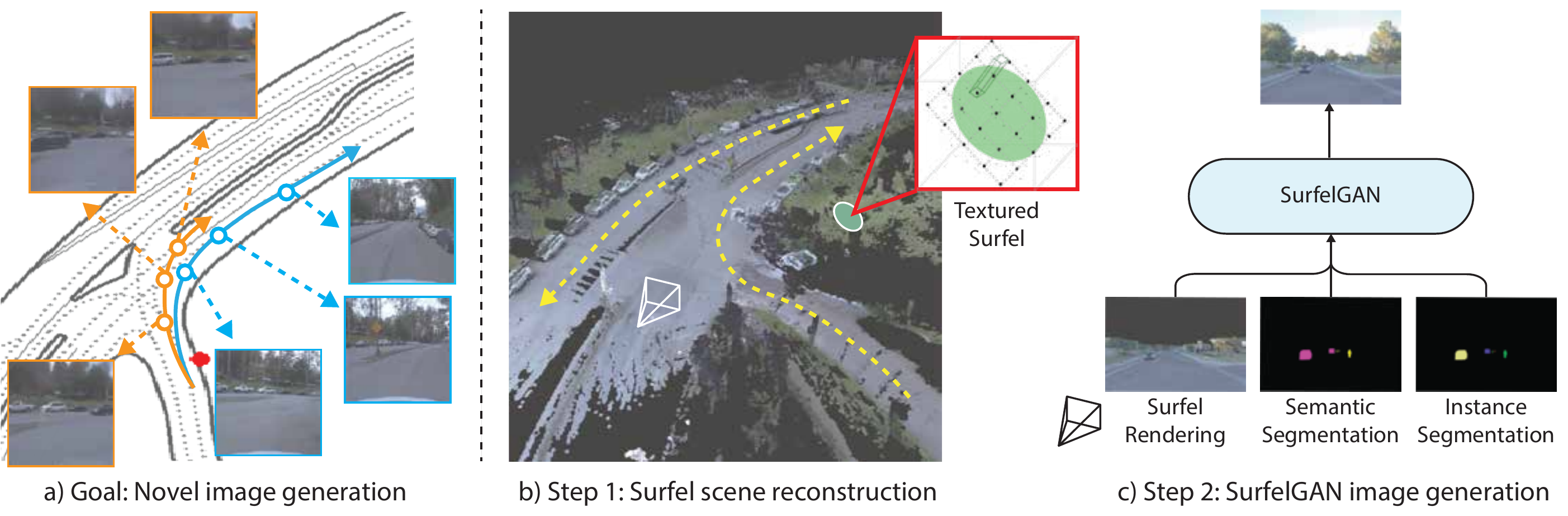}
\caption{\textbf{Overview of our proposed system.}
a) The goal of this work is the generation of camera images for autonomous driving simulation. When provided with a novel trajectory of the self-driving vehicle in simulation, the system generates realistic visual sensor data that is useful for downstream modules such as an object detector, a behavior predictor, or a motion planner. At a high level, the method consists of two steps: b) First, we scan the target environment and reconstruct a scene consisting of rich textured surfels. c) Surfels are rendered at the camera pose of the novel trajectory, alongside semantic and instance segmentation masks. Through a GAN \cite{goodfellow2014generative}, we generate realistically looking camera images.
\vspace{-0.4cm}
}
\label{fig:teaser}
\end{figure*}

Recent advances in deep learning have inspired breakthroughs in multiple areas related to autonomous driving such as perception~\cite{he2017mask, liu2016ssd}, prediction~\cite{bansalchauffeurnet,chai2019multipath} and planning~\cite{everett2018motion}. These recent trends only underscore the increasingly significant role of data-driven system development. One aspect is that deep learning networks benefit from large training datasets. Another is that autonomous driving system evaluation requires the ability to realistically replay a large set of diverse and complex scenarios in simulation capturing sensor properties, seasons, time of day, and weather. Developing simulators that support the levels of realism required for autonomous system evaluation is a challenging task. There are many ways to design simulators, including simulating \textit{mid-level} object representations~\cite{bansalchauffeurnet, dosovitskiy2017carla}. However, mid-level representations omit subtle perceptual cues that are important for scene understanding, such as pedestrian gestures and blinking lights on vehicles. Furthermore, as end-to-end models that combine perception, prediction, and sometimes even control become an increasingly more popular direction of research, we are faced with the need to faithfully simulate the sensor data, which is the input to such models during scenario replay. 

Frameworks for autonomous driving that support realistic sensor simulation are traditionally built on top of gaming engines such as Unreal or Unity~\cite{dosovitskiy2017carla}. The environment and its object models are created and arranged manually, to approximate real-world scenes of interest. In order to enable realistic LiDAR and radar modeling, material properties often need to be manually specified as well. The overall process is time-consuming and effort-intensive. Furthermore, simple ray-casting or ray-tracing techniques are often insufficient to generate realistic camera, LiDAR, or radar data for a specific self-driving system, and additional work is required to adapt the simulated sensor statistics to the real sensors. 

In this work, we propose a simple yet effective data-driven approach for creating realistic scenario sensor data. Our approach relies on camera and LiDAR data collected during a single pass, or several passes, of an autonomous vehicle through a scene of interest. We use this data to reconstruct the scene using a \textit~{texture-mapped surfel} representation. This representation is simple and computationally efficient to create and preserves rich information about the 3D geometry, semantics, and appearance of all objects in the scene. Given the surfel reconstruction, we can render the scene for novel poses of the self-driving vehicle (SDV) and the other scenario agents. The rendered reconstruction for these novel views may have some missing parts due to occlusion differences between the initial and the new scene configuration. It can also have visual quality artifacts due to the limited fidelity of the surfel reconstruction. We address this gap by applying a GAN network~\cite{goodfellow2014generative} to the rendered surfel views to produce the final high-quality image reconstructions. An overview of our proposed system is illustrated in \figref{fig:teaser}.

Our work makes the following contributions: 1) We describe a pipeline that builds a detailed reconstruction of a dynamic scene from real-world sensor data. This representation allows us to render novel views in the scene, corresponding to deviations of the SDV and the other agents in the environment from their initially captured trajectories (Sec.
~\ref{sec:Scene_Reconstruction}). 2) 
We propose a GAN architecture that takes in the rendered surfel views and synthesizes images with quality and statistics approaching that of real images (\tblref{tab:detector}) 3) We build the first dataset for reliably evaluating the task of novel view synthesis for autonomous driving, which contains cases in which two self-driving vehicles observe the same scene at the same time. We use this dataset to provide additional evaluation and demonstrate the usefulness of our SurfelGAN model.

\section{Related Work}
\label{Section:Related:Works}
\PAR{Simulated Environments for Driving Agents.}
 There have been many efforts towards building simulated environments for various tasks~\cite{brodeur2017home,dosovitskiy2017carla,wu2018building, wymann2000torcs,xia2018gibson}. Much work has focused on indoor environments~\cite{brodeur2017home,wu2018building,xia2018gibson} based on public indoor datasets such as SUNCG~\cite{song2017semantic} or Matterport3D~\cite{chang2017matterport3d}. 
 In contrast to indoor settings where the environment is relatively simple and easy to model, simulators for autonomous driving exhibit significant challenges in modeling the complicated and dynamic scenarios of real-world scenes. TORCS\cite{wymann2000torcs} is one of the first simulation environments that support multi-agent racing, but is not tailored for real-world autonomous driving research and development. DeepGTAV~\cite{deep_GTAV_v2} provides a plugin that transforms the Grand Theft Auto gaming environment into a vision-based self-driving car research environment. CARLA\cite{dosovitskiy2017carla} is a popular open-source simulation engine that supports the training and testing of SDVs. All these simulators rely on manual creation of synthetic environments, which is a formidable and laborious process. In CARLA~\cite{dosovitskiy2017carla}, the 3D model of the environment, which includes buildings, road, vegetation, vehicles, and pedestrians, is manually created. The simulator provides one town with 2.9\,km of drivable roads for training and another town with 1.4\,km of drivable roads for testing. In contrast, our system is easily extendable to new scenes that are driven by an SDV. Furthermore, because the environment we are building is a high-quality reconstruction based on the vehicle sensors, it naturally closes the domain gap between synthetic and real contents, which is present in most traditional simulation environments. Similar to this work, AADS~\cite{li2019aads} utilizes real sensor data to synthesize novel views. The major difference is that we reconstruct the 3D~environment, while AADS uses a purely image-based novel view synthesis. Reconstructing the 3D environment gives us the freedom to synthesize novel views that could not be easily captured in the real world. Moreover, once our environment is built, we no longer need to store the images or query the nearest $K$~views upon synthesis, which saves time for deployment. 

\PAR{Learning on Synthetic Data.}
Besides enabling end-to-end training and evaluation of agents, the simulated environment can also provide a large amount of data for training deep neural networks. \cite{ros2016synthia} uses a synthetic scene to generate a large amount of fully labeled training data for urban scene segmentation. \cite{kar2019meta} generate images containing novel placement of dynamic objects to boost the performance of object detection. 

\PAR{Geometric Reconstruction and 3D Representations.} A typical approach for 3D reconstruction of outdoor environments is to use structure from motion~\cite{ullman1979interpretation, wu2011visualsfm} or multi-view stereo~\cite{Furu:2010:PMVS} to recover a dense 3D point cloud from image collections, then optionally use Poisson reconstruction~\cite{kazhdan2006poisson} to obtain a mesh representation. Such a paradigm is most suitable when we have multiple images covering the same area from different perspectives, which is not always true in our case. Thanks to the rapid advancement of LiDAR technology, we can have accurate depth information to complement the camera image data. Our approach leverages the traditional surfel representation~\cite{pfister2000surfels} augmented with fine-grained image textures, which not only greatly simplifies the 3D reconstruction process, but also effectively models object appearance and color with high fidelity. Truncated Signed Distance Functions~\cite{TSDF:1996} and their most recent variants~\cite{DeepSDF:Park_2019_CVPR} are also promising alternatives to surfel-based modeling. Recent work by Aliev et al. \cite{aliev2019neural} that augments 3D cloud points with a learnable neural descriptor for rendering purposes has also shown promising results, however it assumes a static environment and is not applicable in practice to outdoor driving scenarios, which usually contain tens of millions of points.

\PAR{GAN-based Image Translation.}
Generative Adversarial Networks (GAN)~\cite{goodfellow2014generative} have attracted broad interest in both academia and industry. While \cite{goodfellow2014generative} aims to synthesise realistic images directly, \cite{isola2017image} targets the conditional image synthesis setting. Subsequent research~\cite{arjovsky2017wasserstein,karras2017progressive,park2019semantic,zhu2017unpaired} has made great strides in improving the quality of images generated by GAN methods; we refer the readers to~\cite{creswell2018generative} for an overview. \cite{wang2018video} propose a model trained on Cityscapes~\cite{Cordts2016Cityscapes} that can convert videos of semantic segmentation masks into videos of realistic images. Their requirement of having accurate per-pixel semantic annotations of scenes of interest may be difficult to satisfy. In contrast, our approach requires only accurate 3D bounding boxes for the moving objects in the scene, which can be more cost-effective to obtain by human annotation. We believe even this requirement can be further relaxed, by replacing the ground-truth 3D boxes with 3D boxes produced by running a high-quality offline 3D perception pipeline on the SDV sensor data. Finally, in our work we also address the traditional challenge of GAN evaluation by proposing two new metrics that are suitable for the task of novel view synthesis. 
\section{Approach}
\label{Section:Approach}
In this section, we describe the key innovations of this work: 1) texture-enhanced surfel scene reconstruction and 2) SurfelGAN-based image synthesis applied on the novel rendered scene views. Their combination enables the creation of a realistic data-driven sensor simulation environment. 

\subsection{Surfel Scene Reconstruction}
\label{sec:Scene_Reconstruction}

\begin{figure*}
\centering
\footnotesize
\def\imh{0.23\textwidth}
\def\imw{0.23\textwidth}
\newcommand{\T}[1]{\raisebox{-0.5\height}{#1}}
\setlength{\tabcolsep}{1pt}
\begin{tabular}{cccc}

\T{\includegraphics[width=\imw]     {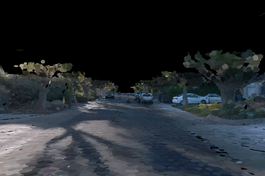}} & 
\T{\includegraphics[width=\imw]      {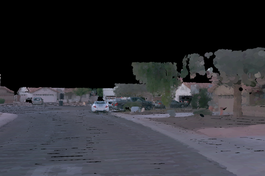}} & 
\T{\includegraphics[width=\imw]    {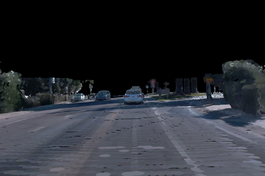}} & 
\T{\includegraphics[width=\imw]      {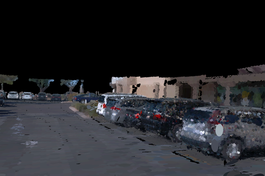}} \\
[+5pt]

\T{\includegraphics[width=\imw]     {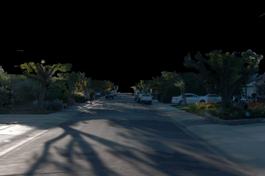}} & 
\T{\includegraphics[width=\imw]      {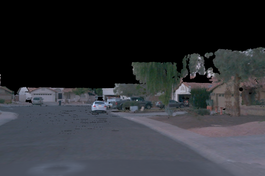}} & 
\T{\includegraphics[width=\imw]    {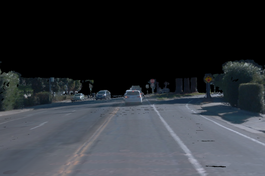}} & 
\T{\includegraphics[width=\imw]      {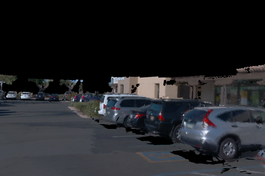}} \\
[+5pt]

\T{\includegraphics[width=\imw]     {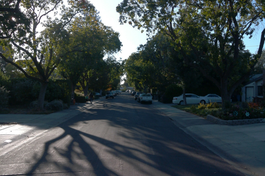}} & 
\T{\includegraphics[width=\imw]      {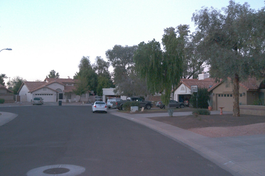}} & 
\T{\includegraphics[width=\imw]    {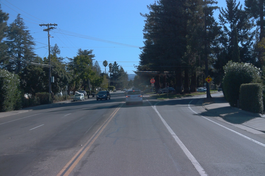}} & 
\T{\includegraphics[width=\imw]      {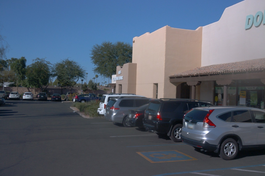}} \\
[+5pt]

\end{tabular}

\caption{Visualization of different scene modeling strategies. \textbf{Top row}: Surfel baseline; \textbf{Center row}: our Texture-Enhanced Surfel Map (also known as \textit{surfel rendering} in the rest of the paper); \textbf{Bottom row}: Real camera image. }
\label{fig:rendering}
\vspace{-0.15in}
\end{figure*}

\PAR{Enhanced Surfel Map.} A good scene reconstruction model enables the faithful preservation of the sensor information, while remaining efficient in terms of computation and storage. Towards this goal, we propose a novel texture-enhanced surfel map representation.  Surfels are compact, easy to reconstruct, and because of their fixed size, easy to texture and compress. Below we describe our approach, which can preserve more fine-grained details compared to traditional surfel map representations~\cite{pfister2000surfels}.   

We discretize the scene into a 3D voxel grid of fixed size and process the LiDAR scans in the order they are captured. For each voxel, we construct a surfel disk by estimating the mean coordinate and the surfel normal, based on all the LiDAR points in that voxel. The surfel disk radius is defined as $\sqrt{3} v$, where $v$ denotes the voxel size. For the LiDAR points binned in a voxel, we also have the corresponding colors from the camera image, which we can use to estimate the surfel color. Note that traditional surfel maps suffer from the trade-off between geometry consistency and fine-grained details, \textit{i.e.,} a large voxel size gives better geometry consistency but fewer details, while small voxel size results in finer details but less stable geometry. Therefore, we take an alternative approach that aims to achieve both good geometry consistency and rich texture details. Specifically, we discretize each surfel disk into a $k\times k$ grid centered on its point centroid, as illustrated in subfigure b) in \figref{fig:teaser}. Each grid center is assigned an independent color to encode higher-resolution texture details. 

Since each surfel may have a different appearance across different frames, due to the variations of the lighting conditions and the changes of relative pose (distance and view angle), we propose to enhance the surfel representation by creating a codebook of such $k\times k$ grids at $n$ various distances. For each grid bin, we determine its color from the first observation, which we found is important to obtain a smooth rendering image. During the rendering stage, we determine which $k\times k$ patch to use based on the camera pose. The final rendering is shown in Fig.~\ref{fig:rendering}. We can see that the baseline surfel map introduces many artifacts at object boundaries and yields non-smooth coloring at non-boundary areas. In contrast, our texture-enhanced surfel map eliminates much of the artifacts and leads to vivid-looking images. In our experiments, we use $v=0.2m$, $k=5$ and $n=10$. 

\PAR{Handling Dynamic Objects.} We consider vehicles as rigid dynamic objects and reconstruct a separate model for each. For simplicity, we leverage the high-quality 3D bounding box annotations from the Waymo Open Dataset~\cite{waymo_open_dataset} to accumulate the LiDAR points from multiple scans for each object of interest. We apply the Iterative Closest Point (ICP)~\cite{ICP:Segal-RSS-09} algorithm to refine the point cloud registration, producing a dense point cloud that allows an accurate, enhanced surfel reconstruction for each vehicle. Please see \supsecref{sec: surfel details} for reconstructed examples. Our approach does not strictly require 3D box ground-truth; we can also leverage state of the art vehicle detection and tracking algorithms~\cite{FaF_2018_CVPR,SOTA_3D_Detector2_2019} to get initial estimates for ICP. However, we leave this experiment for future work. 

When simulating the environment, the reconstructed vehicle models can be placed in any location of choice. In the case of the pedestrians, which are deformable objects, we reconstruct a separate surfel model for each LiDAR scan separately. We allow placement of the reconstructed pedestrian anywhere in the scene for that scan. We leave the task of accurate deformable model reconstruction from multiple scans to future work. 

\begin{figure*}[t!]
\centering
\includegraphics[width=0.9\linewidth]{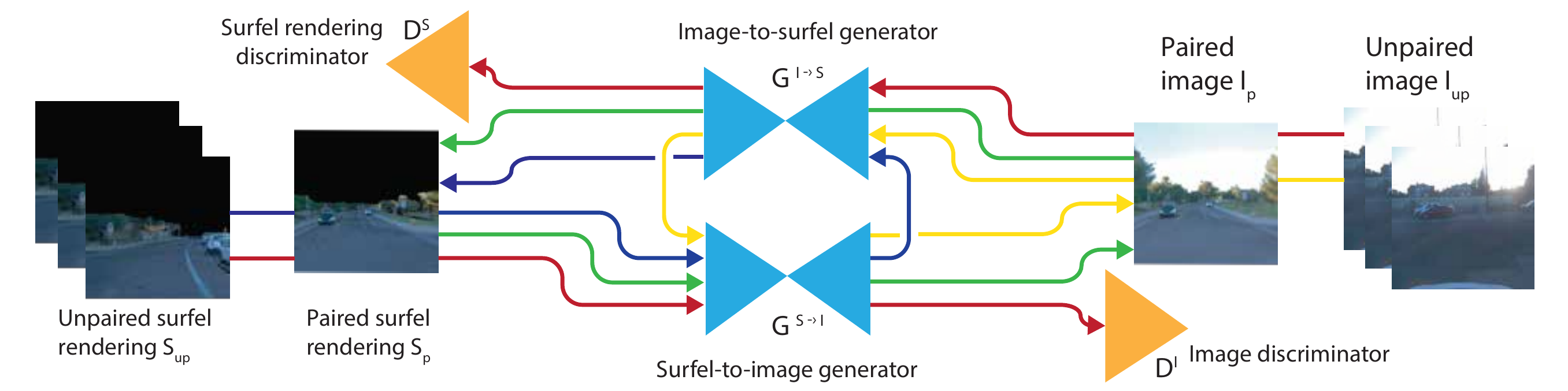}

\caption{(Best viewed in color) \textbf{SurfelGAN training paradigm.} The training setup has two symmetric encoder-decoder generators mapping from surfel renderings to real images $G^{S \rightarrow I}$ and vice versa $G^{I \rightarrow S}$. Additionally, there are two discriminators, $D^{S}$, $D^I$, which specialize in the surfel and the real domain. The losses are shown as colored arrows. Green: supervised reconstruction loss. Red: adversarial loss. Blue/Yellow: cycle-consistency losses. When training with paired data, \eg WOD-TRAIN, the surfel renderings translate to real images, and we can apply a one-directional supervised reconstruction loss (SurfelGAN-S) only or add an additional adversarial loss (SurfelGAN-SA). When training with unpaired data, we can go either from the surfel renderings (\eg WOD-TRAIN-NV) or the real images (\eg Internal Camera Dataset), use one of the encoder-decoder networks to get to the other domain and back. We can then apply a cycle consistency loss. (SurfelGAN-SAC). The encoder-decoder networks consist of 8 convolutional and 8 deconvolutional layers. Discriminators consist of 5 convolutional layers. All network operate on $256\times256$ sized input.\vspace{-0.4cm}}
\label{fig:networks}
\end{figure*}

\subsection{Image Synthesis via SurfelGAN}
\label{sec:Image_Synthesis}

While the surfel scene reconstruction provides a rich representation of the environment, it produces surfel-based renderings that have a non-negligible realism gap when compared to real images, due to incomplete reconstruction and imperfect geometry and texturing (see \figref{fig:rendering}). Our SurfelGAN model is explicitly designed to address this issue. 

SurfelGAN is a generative model that converts surfel  image renderings to realistically looking images. We treat semantic and instance segmentation maps as additional rendered image channels. For the sake of simplicity, we omit their explicit mention in the rest of this this section.

Let the generator $G^{S \rightarrow I}_{\theta_S}$ be an encoder-decoder model with learnable parameters $\theta_S$. Given pairs of surfel renderings $\setS_p$ and images $\setI_p$, the supervised loss can be applied to train the generator. We call a SurfelGAN model that is trained solely with supervised learning \textbf{SurfelGAN-S}. Additionally, we can add an adversarial loss from a real image discriminator $D^{I}_{\phi_I}$. SurfelGANs trained with this additional loss is named \textbf{SurfelGAN-SA}.

However, paired training data between surfel renderings and real image is very limited. Unpaired data is however easy to obtain. We leverage unpaired data for two purposes: improving the generalization of the discriminator by training with more unlabeled examples, and regularizing the generator by enforcing cycle consistency. Let the reverse generator $G^{I \rightarrow S}_{\theta_I}$ be another encoder-decoder model which has the same architecture as $G^{S \rightarrow I}_{\theta_S}$ except more output channels for semantic and instance maps. Then \textit{any} surfel rendering, paired $\setS_p$ or unpaired $\setS_{u}$ can be translated to a real image and translated back to a surfel rendering, where a cycle consistency loss can be applied. The same applies to \textit{any} paired $\setI_p$ or unpaired $\setI_{u}$ real image as well. Finally, we add the surfel rendering discriminator $D^{S}_{\phi_S}$ that judges generated surfel images. We call SurfelGANs trained with additional cycle consistency \textbf{SurfelGAN-SAC}. An intuitive overview of the training strategy is shown in \figref{fig:networks}, while \secref{Section:Experimental Results} contains a detailed description of our paired and unpaired data. We optimize the following objective: 
\begin{equation} \label{eq1}
\small{
\begin{split}
&\max_{\phi_{S}, \phi_{I}}\min_{\theta_{S}, \theta_{I}} 
\mathcal{L}_r(G^{S \rightarrow I}_{\theta_S}, \setS_p, \setI_p) + \lambda_1\mathcal{L}_r(G^{I \rightarrow S}_{\theta_I}, \setI_p, \setS_p) \\
& + \lambda_2\mathcal{L}_{a}(G^{S \rightarrow I}_{\theta_S}, D^{I}_{\phi_I}, \setS_{p, u} )+\lambda_3\mathcal{L}_{a}(G^{I \rightarrow S}_{\theta_I}, D^{S}_{\phi_S}, \setI_{p, u}) \\
& + \lambda_4\mathcal{L}_{c}(G^{S \rightarrow I}_{\theta_S}, G^{I \rightarrow S}_{\theta_I}, \setS_{p, u}) 
 + \lambda_5\mathcal{L}_{c}(G^{I \rightarrow S}_{\theta_I}, G^{S \rightarrow I}_{\theta_S}, \setI_{p, u}),
\end{split}
}
\end{equation}
where $\mathcal{L}_{r}$, $\mathcal{L}_{a}$, $\mathcal{L}_{c}$ denote the supervised reconstruction, adversarial and cycle consistency loss, respectively. We use hinged Wasserstein loss for adversarial training~\cite{lim2017geometric,miyato2018spectral,zhang2018self} in our experiments as it helps to stabilize the training. We use \lone-loss as reconstruction and cycle-consistency loss for renderings and images and cross entropy loss for semantic and instance maps. 

\PAR{Distance Weighted Loss.}
Due to the limited coverage of the surfel map, our surfel rendering contains large areas of unknown regions. The uncertainty in those regions is much higher than that of the region with surfel information. Also, the distance between the camera and the surfel introduces another factor of uncertainty. Therefore, we use a distance weighted loss to stabilize our GAN training. Specifically, during data pre-processing, we generate a distance map that records the nearest distance to the observed region and then uses the distance information as weighting coefficients to modulate our reconstruction loss. 

\PAR{Training Details.}
We use the Adam\cite{kingma2014adam} optimizer for training. We set the initial learning rate to $2\mathrm{e}{-4}$ for both the generator and the discriminator and set $\beta_1=0.5$ and $\beta_2=0.9$. We use batch normalization \cite{ioffe2015batch} after Relu activation. We set $\lambda_1=1,\lambda_2,\lambda_3=0.001, ,\lambda_4,\lambda_5=0.1$ in all of our experiments. The total training time of our network is 3 days, based on one Nvidia Titan V100 GPU with batch size 8.

\section{Experimental Results}
\label{Section:Experimental Results}

\begin{figure*}
\centering
\footnotesize
\def\imh{0.15\textwidth}
\def\imw{0.15\textwidth}
\newcommand{\T}[1]{\raisebox{-0.5\height}{#1}}
\setlength{\tabcolsep}{2pt}
\begin{tabular}{ccccccc}

\rotatebox[origin=c]{90}{Surfel Rendering} &
\T{\includegraphics[width=\imw]     {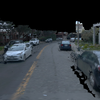}} & 
\T{\includegraphics[width=\imw]   {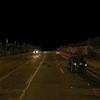}} & 
\T{\includegraphics[width=\imw]      {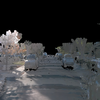}} & 
\T{\includegraphics[width=\imw]    {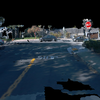}} & 
\T{\includegraphics[width=\imw]    {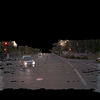}} & 
\T{\includegraphics[width=\imw]     
{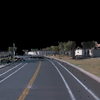}} \\

\rotatebox[origin=c]{90}{SurfelGAN-S} &
\T{\includegraphics[width=\imw]     {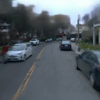}} & 
\T{\includegraphics[width=\imw]   {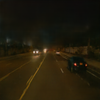}} & 
\T{\includegraphics[width=\imw]      {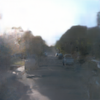}} & 
\T{\includegraphics[width=\imw]    {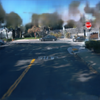}} & 
\T{\includegraphics[width=\imw]    {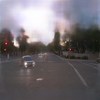}} & 
\T{\includegraphics[width=\imw]      {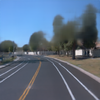}} \\

\rotatebox[origin=c]{90}{SurfelGAN-SA} &
\T{\includegraphics[width=\imw]     {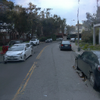}} & 
\T{\includegraphics[width=\imw]   {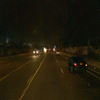}} & 
\T{\includegraphics[width=\imw]      {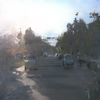}} & 
\T{\includegraphics[width=\imw]    {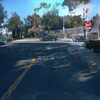}} & 
\T{\includegraphics[width=\imw]    {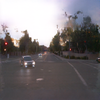}} & 
\T{\includegraphics[width=\imw]      {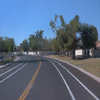}} \\

\rotatebox[origin=c]{90}{SurfelGAN-SAC} &
\T{\includegraphics[width=\imw]     {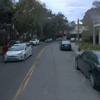}} & 
\T{\includegraphics[width=\imw]   {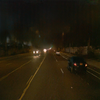}} & 
\T{\includegraphics[width=\imw]      {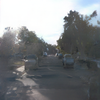}} & 
\T{\includegraphics[width=\imw]    {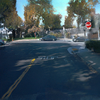}} & 
\T{\includegraphics[width=\imw]    {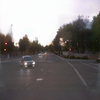}} & 
\T{\includegraphics[width=\imw]      {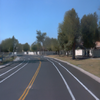}} \\

\rotatebox[origin=c]{90}{REAL} &
\T{\includegraphics[width=\imw]     {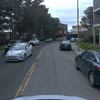}} & 
\T{\includegraphics[width=\imw]   {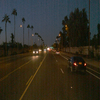}} & 
\T{\includegraphics[width=\imw]      {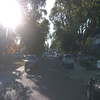}} & 
\T{\includegraphics[width=\imw]    {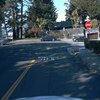}} & 
\T{\includegraphics[width=\imw]    {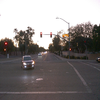}} & 
\T{\includegraphics[width=\imw]      {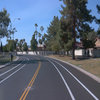}} \\

\end{tabular}
\caption{Qualitative comparison bewteen different SurfelGAN variants and the baseline on WOD-EVAL under different weather conditions.}
\label{fig:qualitative}
\vspace{-0.15in}
\end{figure*}

We base our experiments mainly on the Waymo Open Dataset \cite{waymo_open_dataset}, but we also collected two additional datasets in order to obtain a higher quality model and enable a more extensive evaluation.

\PAR{Waymo Open Dataset (WOD) \cite{waymo_open_dataset}.} The dataset consists of 798 training (\textbf{WOD-TRAIN}) and 202 validation (\textbf{WOD-EVAL}) sequences. Each sequence contains 20 seconds of camera and LiDAR data captured at 10Hz, as well as fully annotated 3D bounding boxes for vehicles, pedestrians, and cyclists. The LiDAR data covers a full 360 degrees around the agent, while five cameras capture the frontal 180 degrees. After reconstructing the surfel scenes, we can render the surfel images in the same pose as the original camera images, hence generating surfel-image-to-camera-image pairs that can be used for paired training and evaluation.
Since during the reconstruction process we know the category for each surfel, we can easily derive both semantic and instance segmentation masks by first rendering an index map that associates each pixel with a surfel index and then determining the semantic class or instance number through a look-up table.

We derive another dataset from WOD, which we call Waymo Open Dataset-Novel View (\textbf{WOD-TRAIN-NV} and \textbf{WOD-EVAL-NV}). We again start from reconstructed surfel scenes, but we now render surfel images from novel camera poses perturbed from existing camera poses. The perturbation consists of applying a random translation and a random yaw angle perturbation to the camera mounted vehicle. We use the annotated 3D bounding boxes to ensure the perturbed vehicle does not intersect with other objects in the scene.

We generate one new surfel image rendering for each frame in the original dataset. Note that although this dataset comes for free, \ie we could generate any number of testing frames, it does not have corresponding camera images. Therefore, this dataset can only be used for unpaired training and only some types of evaluation.

\PAR{Internal Camera Image Dataset.} We collected additional 9.8k short sequences (100 frames for each) similar to WOD images. These un-annotated images are used for unpaired training of real images.

\PAR{Dual-Camera-Pose Dataset (DCP)} Finally, we built a unique dataset tailored for measuring the realism of our model. The dataset contains scenarios where two vehicles observe the same scene at the same time. Specifically, we find the interval where two vehicles are within $20m$ of each other. We use the sensor data from the first vehicle to reconstruct the scene, and render the surfel image at the exact pose of the second vehicle. After filtering cases where the scene reconstruction is too incomplete, we obtain around 1k pairs, for which we can directly measure the pixel-wise accuracy of the generated image.

\begin{table*}[!tbp]
  \footnotesize
  \centering
  \setlength{\tabcolsep}{5.4pt}
  \begin{tabular}{l|cccc|cccc|cccc}
\toprule
  & \multicolumn{4}{c|}{WOD-TRAIN-NV} & \multicolumn{4}{c|}{WOD-EVAL} & \multicolumn{4}{c}{WOD-EVAL-NV} \\
  
  & AP@50 $\uparrow$ & AP@75 $\uparrow$ & AP $\uparrow$ & Rec $\uparrow$ & AP@50 & AP@75 & AP & Rec & AP@50 & AP@75 & AP & Rec \\
\midrule  
Surfel (baseline) & 0.444 & 0.168 & 0.211 & 0.342 & 0.521 & 0.168 & 0.239 & 0.371 & 0.462 & 0.154 & 0.213 &0.348 \\
  
SurfelGAN-S (ours) & 0.508 & 0.177 & 0.236 & 0.359 & 0.576 & 0.164  &0.252 & 0.341 & 0.514 & 0.159 & 0.230 & 0.358 \\
   
SurfelGAN-SA (ours) & 0.554 & 0.200 & 0.259 & 0.382 & 0.610 & 0.174 & 0.266 & 0.394 & 0.567 & 0.180 & 0.257 & 0.387 \\

SurfelGAN-SAC (ours) & \textbf{0.564} & \textbf{0.200} & \textbf{ 0.263 } & \textbf{0.385} & \textbf{0.620} & \textbf{0.181} & \textbf{0.272} &\textbf{0.400} & \textbf{0.570} & \textbf{0.181} & \textbf{0.258} & \textbf{0.388} \\

\midrule  

Real (upper bound) & - & - & - & - & 0.619 & 0.198 & 0.281& 0.424 & - & - & - & - \\
\bottomrule
  \end{tabular}
\caption{ Realism \wrt an off-the-shelf vehicle object detector. We generated images using the proposed SurfelGAN and ran inference on them using an off-the-shelf object detector. We report the standard COCO object detection metrics \cite{Lin14}, including variants of the average-precision (AP) and recall at 100 (Rec). \textbf{ Surfel} is the surfel rendering that is the input to SurfelGAN. \textbf{SurfelGAN} is the proposed model. The S variant is trained with paired supervised learning only. The SA variant adds the adversarial loss, and the SAC variant makes use of additional unpaired data and applied a cyclic adversarial loss. \textbf{Real} is the real image captured by cameras, which is only available in WOD-EVAL. It serves as an upper bound to the detector's quality. As shown above, SurfelGAN significantly improves over the baseline, and reaches quality metric values similar to those of the real images.}
\label{tab:detector}
\end{table*}

\subsection{Model Variants and Baseline}

Most experiments were performed on three variants of our proposed model. \textbf{Supervised (S)}: we train the surfel-rendering-to-image model in a supervised way by minimizing an \lone-loss between the generated image and the ground-truth real image. This type of training requires paired data. Hence, it is only possible to train on WOD-TRAIN. \textbf{Supervised + Adversarial (SA)}: we still only consider WOD-TRAIN as the training data. However, we add an adversarial loss alongside the \lone-loss. \textbf{Supervised + Adversarial + Cycle (SAC)}: in this variation, we also use WOD-TRAIN-NV and the Internal Camera Image Dataset. Since these two sets are unpaired, the supervised loss does not apply. We propose to use a cycle-consistency loss in addition to the adversarial loss, as discussed in \secref{sec:Image_Synthesis}.

The baseline for our applications is the direct surfel renderings (\textbf{Surfel}) that serve as the input to our model.

\subsection{Vehicle Detector Realism}

Since the primary application of this work is simulation for autonomous driving, it is natural to evaluate the generated camera data using a downstream perception module. Specifically, we want to know how well an off-the-shelf vehicle object detector performs on the generated images without any fine-tuning. This is a test of whether the detector statistics on the generated images match those it obtains on the real images. We chose to use a vehicle detector with a ResNet architecture \cite{he2016deep} and an SSD detection head \cite{liu2016ssd}, trained and evaluated on resized images in $512\times512$ resolution from a mixture of datasets that include WOD-TRAIN.

We trained our SurfelGAN model variants on a mixture of WOD-TRAIN, WOD-TRAIN-NV and the Internal Camera Image Dataset, and generated images on WOD-TRAIN-NV, WOD-EVAL and WOD-EVAL-NV. \tblref{tab:detector} shows the quantitative comparison of the detector's quality on the original surfel renderings that are the input of SurfelGAN, on images generated from the variants of SurfelGAN and on real images. A few generated image examples are displayed in \figref{fig:qualitative}. Please find additional visualizations in the supplementary material. \figref{fig:novelview} highlights our system's ability to generate images in novel views.

Notably, our texture-enhanced surfel scene reconstruction already produces surfel renderings that achieve good detection quality on the WOD-EVAL set at 52.1\% AP@50. But there is still a significant gap between these surfel renderings and real images at 61.9\%, which motivates our SurfelGAN work. As shown in \tblref{tab:detector}, SurfelGAN-S, -SA and -SAC variants gradually improve over the baseline surfel renderings. SurfelGAN-SAC ultimately improves the AP@50 metric from 52.1\% to 62.0\% on WOD-EVAL, which is on par with the real images at 61.9\%. This shows that images generated by SurfelGAN-SAC are close to real images in the eyes of the detector, which is the primary motivation of this work.

There are two types of generalization worth evaluating. The first type is whether a SurfelGAN model trained on one set of scenes (\eg WOD-TRAIN and WOD-TRAIN-NV) generalizes to new scenes (\eg WOD-EVAL and WOD-EVAL-NV). We believe that the SurfelGAN model generalizes well since the relative improvement of SurfelGAN over the baseline is very similar between the WOD-TRAIN-NV and WOD-EVAL-NV columns in \tblref{tab:detector}.

The second type of generalization is whether surfel rendering has a strong bias towards the poses from which the scene was reconstructed. We compare the metric values between WOD-EVAL and WOD-EVAL-NV. Although SurfelGAN improved by roughly 10\% over the baseline in both cases, there is a noticeable quality difference between the two columns. To better understand this difference, in \tblref{tab:breakdown}, we breakdown the metrics of SurfelGAN-SAC on WOD-EVAL-NV according to how much each pose deviates from the original poses in WOD-EVAL. The deviation $d(.)$ is defined as a weighted sum of both translational and rotational differences of the poses:
\begin{equation}
    d((t, R),(t', R')) = ||t - t'|| + \lambda_R \frac{|\log(R^TR')||}{\sqrt{2}}
\end{equation}
where $t$ and $R$ are the pose (translation and rotation) of the novel view in WOD-EVAL-NV, and $t'$, $R'$ the pose of its closest pose in WOD-EVAL. $\lambda_R$ is chosen to be $1.0$. The result suggests that the surfel renderings do have a quality bias w.r.t viewing direction, which means we should not perturb too much from the original poses if we want higher quality synthesized data. However, we believe that this problem can be ameliorated if we were to reconstruct the surfel scene from multiple runs, a direction we left to future work. 

\begin{table}[!tbp]
  \footnotesize
  \centering
  \begin{tabular}{lcccc}
  \toprule
  Perturbation & AP@50 & AP@75 & AP\\
  \hline
   $d <= 1.0$ & 0.574 & 0.174 & 0.257\\
   $1.0 < d <= 2.0$  & 0.547 & 0.173 & 0.246\\
   $2.0 < d$  & 0.488 & 0.153 & 0.218\\
  \bottomrule
  \end{tabular}
 \caption{Detector metric break down at different perturbation levels on WOD-EVAL-NV by SurfelGAN-SAC.}
 \label{tab:breakdown}
\end{table}

\subsection{Image-Pixels Realism}

\begin{table}[!tbp]
  \footnotesize
  \centering
  \begin{tabular}{lcccc}
\toprule
&Surfel & SGAN-S & SGAN-SA & SGAN-SAC \\
\midrule  
\lone-distance $\downarrow$ &0.262 & \textbf{0.229} & 0.240 & 0.238 \\
\bottomrule
  \end{tabular}
\caption{Image-pixel realism. We applied the SurfelGAN on the Dual-Camera-Pose Dataset, where it is possible to measure \lone-distance error between the generated images and the real ones.}
\label{tab:pixel}
\end{table}

The Dual-Camera-Pose (DCP) Dataset contains scenarios in which two vehicles observe the same scene at the same time, allowing us to can reconstruct the surfel scene using one camera and generate images from the point of view of the second camera. We can match each generated image to the real one and report the \lone-distance error on the pixels that are covered by the surfel rendering. This is to ensure that there is a fair comparison between the surfel renderings and the generated images. Like in the previous experiment, the model is trained using WOD-TRAIN, WOD-TRAIN-NV, and the Internal Camera Image Dataset. The results are shown in \tblref{tab:pixel}. SurfelGAN improves on top of the surfel renderings, generating images that are closer to real images in \lone-distance. However, it is worth noting that the SurfelGAN-S version outperforms both SA and SAC that used additional losses and data during training. This finding is not unexpected since SurfelGAN-S optimizes for the \lone-distance.

\begin{figure}
\centering
\footnotesize
\def\imh{0.15\textwidth}
\def\imw{0.15\textwidth}
\newcommand{\T}[1]{\raisebox{-0.5\height}{#1}}
\setlength{\tabcolsep}{1pt}
\begin{tabular}{ccc}

\T{\includegraphics[width=\imw]     {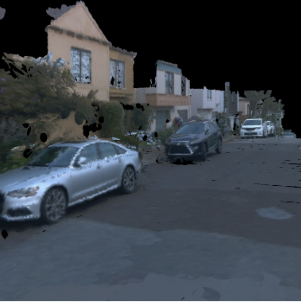}} & 
\T{\includegraphics[width=\imw]      {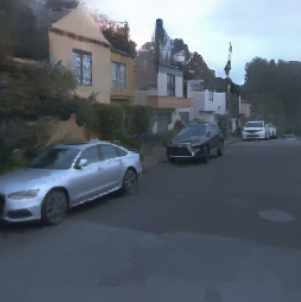}} & 
\T{\includegraphics[width=\imw]    {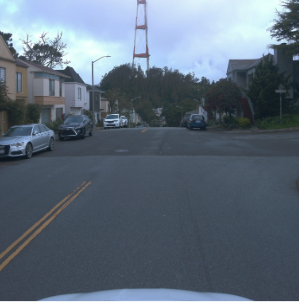}} \\

\T{\includegraphics[width=\imw]     {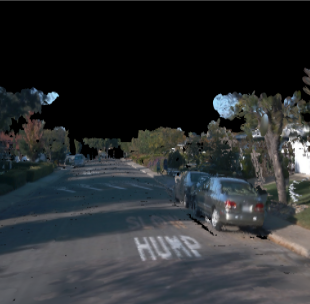}} & 
\T{\includegraphics[width=\imw]      {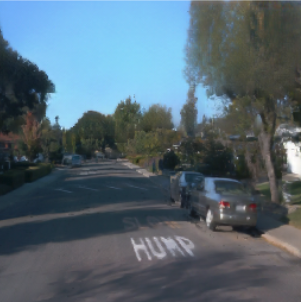}} & 
\T{\includegraphics[width=\imw]    {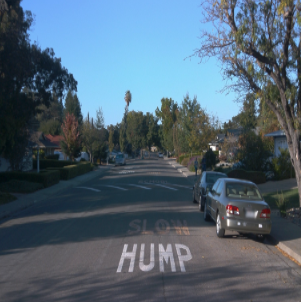}}


\end{tabular}

\caption{\small{Novel View Synthesis. The first column is Surfel image under novel view, the second column is our synthesized result. The third column is the original view. Additional visualization can be found in \supsecref{Section:supp}.}}
\label{fig:novelview}
\vspace{-0.15in}
\end{figure}
\subsection{Improving Detection by Data Augmentation}
In an additional experiment, we explore whether the SurfelGAN-generated images from perturbed views are a helpful form of data augmentation for training a vehicle object detector. 
 For the baseline, we trained a vehicle detector on WOD-TRAIN and evaluated the detector's quality on WOD-EVAL. We then trained another vehicle detector using both WOD-TRAIN and surfel images generated from WOD-TRAIN-NV, and also evaluated on WOD-EVAL.

WOD-TRAIN-NV only inherits 3D bounding boxes from WOD-TRAIN, and does not contain tightly-fitting 2D bounding boxes like those in WOD-TRAIN. We approximate the latter by projecting all surfels in the 3D bounding boxes to the 2D novel view and taking the axis-aligned bounding box as an approximation.
The results are shown in \tblref{tab:augmentation}. The data augmentation significantly boosts the average precision metric, improving the AP@50 score from 21.9\% to 25.4\%, the AP@75 from 10.8\% to 12.1\%, and the average AP from 11.9\% to 13.0\%. It is worth noting that these AP scores are much lower than those in \tblref{tab:detector}. The main reason for the discrepancy is that images are resized differently in order to use the off-the-shelf detector in \tblref{tab:detector}. 
We also trained on surfel renderings directly. There is a slight improvement using them compared to training only on WOD-TRAIN. Using SurfelGAN synthesized images yields a much more significant improvement, which further demonstrates the realism of the SurfelGAN model. 

\begin{table}[!tbp]
  \footnotesize
  \centering
  \begin{tabular}{lcccc}
  \toprule
  Training Set & AP@50 & AP@75 & AP \\
  \hline
   WOD-TRAIN & 0.219 & 0.108 & 0.119\\
   + WOD-TRAIN-NV Surfel & 0.228 & 0.111 & 0.120\\
   + WOD-TRAIN-NV SurfelGAN  & \textbf{0.254} & \textbf{0.121} & \textbf{0.130}\\
  \bottomrule
  \end{tabular}
 \caption{Detector metric on Open Dataset validation set when trained with different combination of data.}
 \label{tab:augmentation}
\end{table}

\subsection{Limitations}

\begin{figure}
\centering
\footnotesize
\def\imh{0.15\textwidth}
\def\imw{0.15\textwidth}
\newcommand{\T}[1]{\raisebox{-0.5\height}{#1}}
\setlength{\tabcolsep}{1pt}
\begin{tabular}{ccc}

\T{\includegraphics[width=\imw]     {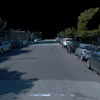}} & 
\T{\includegraphics[width=\imw]      {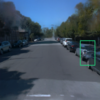}} & 
\T{\includegraphics[width=\imw]    {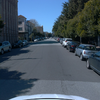}} \\

\T{\includegraphics[width=\imw]     {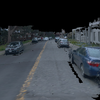}} & 
\T{\includegraphics[width=\imw]      {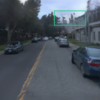}} & 
\T{\includegraphics[width=\imw]    {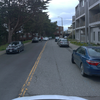}}


\end{tabular}

\caption{\small{We show two typical failure cases, the first is the case where reconstructed Surfel map contains too large errors. The second is on the unmapped region (top part of building in this example). }}
\label{fig:failure}
\vspace{-0.15in}
\end{figure}
The surfel scene reconstruction has its limitations. For example, it may fail to reconstruct certain areas of the scene, as illustrated in the top row in \figref{fig:failure}. In this particular case, SurfelGAN was unable to recover from broken geometry, resulting in an unrealistically looking vehicle. Suboptimal renderings also may occur in areas that are missing surfels.  Lacking any surfel cues forces the model output to have high variance, especially when it tries to hallucinate patterns that infrequently appear in the dataset, such as tall buildings. This observation suggests that there is room for improvement in the reconstruction stage. In the case where we only have partial geometry, applying a learned geometry completion model first could be more helpful than relying solely on the generating module to resolve all artifacts.

\section{Conclusion}
\label{Section:Conclusions}
We propose a simple yet effective data-driven approach, which can synthesize camera data for autonomous driving simulations. Based on the camera and LiDAR data captured by a vehicle pass through a scene, we reconstruct a 3D model using our Enhanced Surfel Map representation. Given this representation, we can render novel views and configurations of objects in the environment. We use our SurfelGAN image synthesis model to fix any reconstruction, occlusion or rendering artifacts. To the best of our knowledge, we have built the first purely data-driven camera simulation system for autonomous driving. Experimental results not only demonstrate the high level of realism of our synthesized sensor data but also show the data can be used for training dataset augmentation for deep neural networks. In future work, we plan to enhance camera simulation further by improving the dynamic object modeling process and by investigating temporally consistent video generation. 

{\small
\bibliographystyle{ieee_fullname}
\bibliography{references}
}

\clearpage

\appendix

\section{Additional Details on Surfel Scene Reconstruction}
\label{sec: surfel details}
\begin{figure}[H]
\centering
\includegraphics[width=0.5\textwidth]{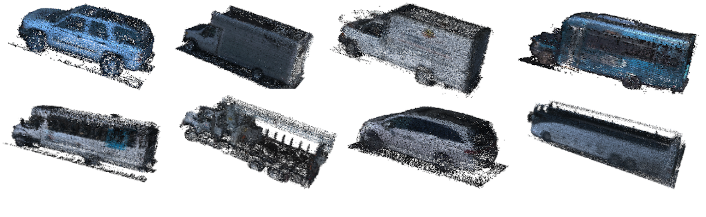}
\caption{Sample reconstructed surfel vehicles.}
\label{fig:supp_vehicle}
\vspace{-0.15in}
\end{figure}

\begin{figure}[H]
\centering
\includegraphics[width=0.5\textwidth]     {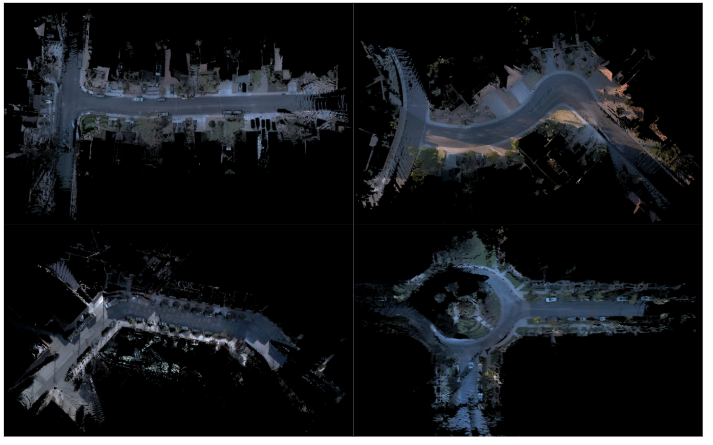}
\caption{Reconstructed surfel scene maps from four different scenes.}
\label{fig:supp_surfel_map}
\vspace{-0.15in}
\end{figure}

For vehicle reconstruction, we aggregate LiDAR points across frames into a single model using bounding box annotations and a local registration method (ICP). We also exploit the symmetry of man-made objects to complete their geometry. We reconstruct a total of 46,786 vehicle models. Some examples are shown in Figure \ref{fig:supp_vehicle}. We also include some surfel scene reconstruction results in Figure \ref{fig:supp_surfel_map}. These surfel maps are built based on camera-LiDAR sequences from the Waymo Open Dataset training split. 

\section{Additional Qualitative Synthesized Image Results}
\label{Section:supp}
\begin{figure*}
\centering
\footnotesize
\def\imh{0.15\textwidth}
\def\imw{0.15\textwidth}
\newcommand{\T}[1]{\raisebox{-0.5\height}{#1}}
\setlength{\tabcolsep}{1pt}
\begin{tabular}{cccccc}

\T{\includegraphics[width=\imw]     {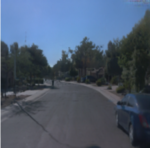}} & 
\T{\includegraphics[width=\imw]   {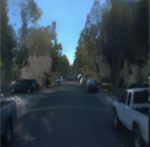}} & 
\T{\includegraphics[width=\imw]      {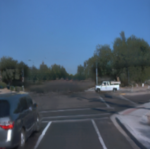}} & 
\T{\includegraphics[width=\imw]    {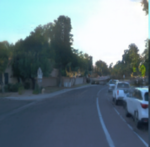}} & 
\T{\includegraphics[width=\imw]      {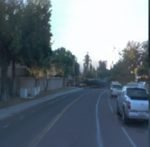}} & 
\T{\includegraphics[width=\imw]    {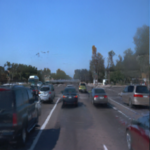}} \\
[+21pt]
\T{\includegraphics[width=\imw]     {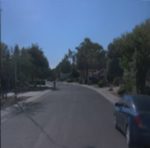}} & 
\T{\includegraphics[width=\imw]   {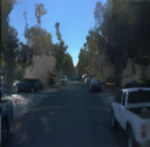}} & 
\T{\includegraphics[width=\imw]      {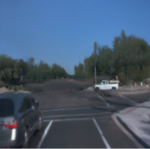}} & 
\T{\includegraphics[width=\imw]    {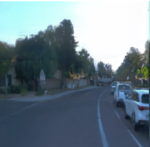}} & 
\T{\includegraphics[width=\imw]      {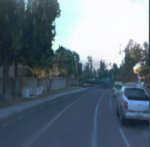}} & 
\T{\includegraphics[width=\imw]    {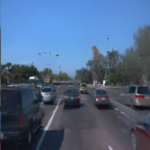}} \\
[+21pt]
\T{\includegraphics[width=\imw]     {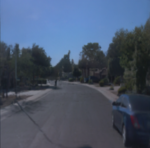}} & 
\T{\includegraphics[width=\imw]   {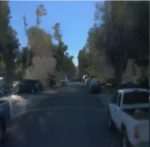}} & 
\T{\includegraphics[width=\imw]      {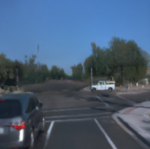}} & 
\T{\includegraphics[width=\imw]    {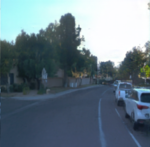}} & 
\T{\includegraphics[width=\imw]      {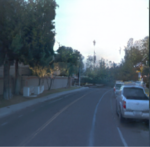}} & 
\T{\includegraphics[width=\imw]    {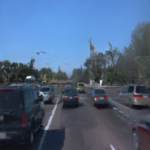}} \\
[+21pt]
\T{\includegraphics[width=\imw]     {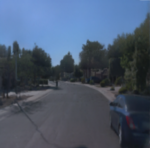}} & 
\T{\includegraphics[width=\imw]   {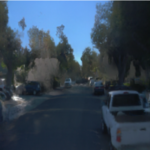}} & 
\T{\includegraphics[width=\imw]      {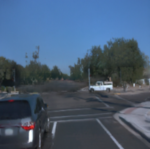}} & 
\T{\includegraphics[width=\imw]    {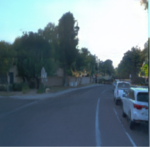}} & 
\T{\includegraphics[width=\imw]      {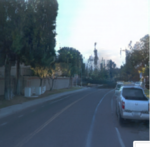}} & 
\T{\includegraphics[width=\imw]    {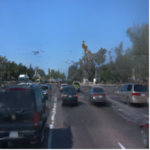}} \\
[+21pt]

\end{tabular}
\caption{Synthesized images when gradually perturbing the camera view point. Each column contains one example. }
\label{fig:supp4}
\vspace{-0.15in}
\end{figure*}

\begin{figure*}
\centering
\footnotesize
\def\imh{0.15\textwidth}
\def\imw{0.15\textwidth}
\newcommand{\T}[1]{\raisebox{-0.5\height}{#1}}
\setlength{\tabcolsep}{1pt}
\begin{tabular}{cccccc}

\T{\includegraphics[width=\imw]     {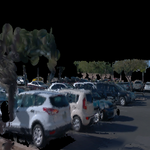}} & 
\T{\includegraphics[width=\imw]   {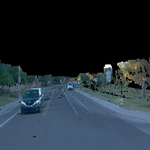}} & 
\T{\includegraphics[width=\imw]      {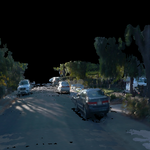}} & 
\T{\includegraphics[width=\imw]    {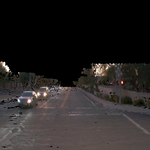}} & 
\T{\includegraphics[width=\imw]      {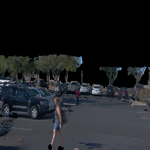}} & 
\T{\includegraphics[width=\imw]    {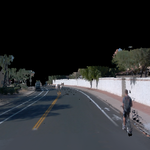}} \\

\T{\includegraphics[width=\imw]     {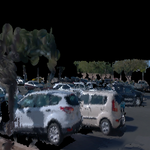}} & 
\T{\includegraphics[width=\imw]   {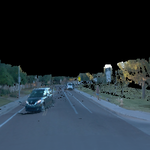}} & 
\T{\includegraphics[width=\imw]      {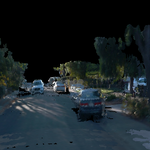}} & 
\T{\includegraphics[width=\imw]    {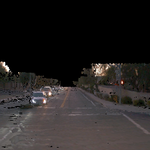}} & 
\T{\includegraphics[width=\imw]      {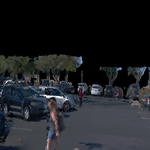}} & 
\T{\includegraphics[width=\imw]    {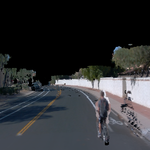}} \\

\T{\includegraphics[width=\imw]     {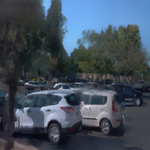}} & 
\T{\includegraphics[width=\imw]   {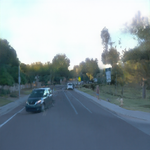}} & 
\T{\includegraphics[width=\imw]      {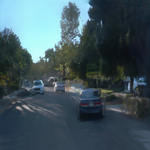}} & 
\T{\includegraphics[width=\imw]    {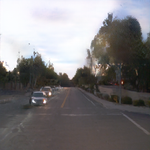}} & 
\T{\includegraphics[width=\imw]      {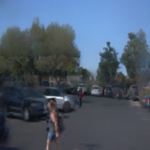}} & 
\T{\includegraphics[width=\imw]    {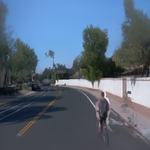}} \\

\end{tabular}
\caption{Synthesized images by perturbing objects in the scene. Top row: original surfel rendering. Center row: surfel rendering after perturbing the objects. Third row: synthesized images. }
\label{fig:supp5}
\vspace{-0.15in}
\end{figure*}

\begin{figure*}
\centering
\footnotesize
\def\imh{0.15\textwidth}
\def\imw{0.15\textwidth}
\newcommand{\T}[1]{\raisebox{-0.5\height}{#1}}
\setlength{\tabcolsep}{1pt}
\begin{tabular}{cccccc}

\T{\includegraphics[width=\imw]     {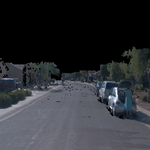}} & 
\T{\includegraphics[width=\imw]   {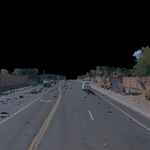}} & 
\T{\includegraphics[width=\imw]      {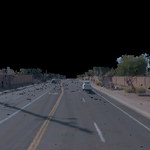}} & 
\T{\includegraphics[width=\imw]    {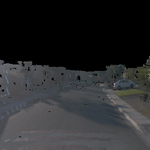}} & 
\T{\includegraphics[width=\imw]      {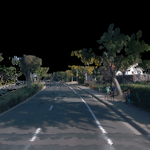}} & 
\T{\includegraphics[width=\imw]    {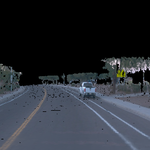}} \\
[+21pt]
\T{\includegraphics[width=\imw]     {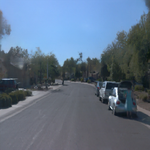}} & 
\T{\includegraphics[width=\imw]   {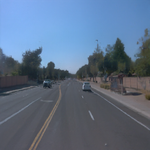}} & 
\T{\includegraphics[width=\imw]      {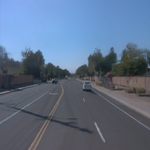}} & 
\T{\includegraphics[width=\imw]    {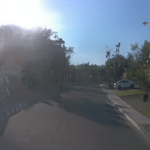}} & 
\T{\includegraphics[width=\imw]      {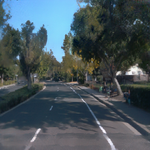}} & 
\T{\includegraphics[width=\imw]    {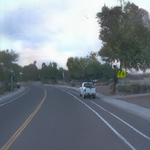}} \\
[+21pt]
&&&&\\

\T{\includegraphics[width=\imw]     {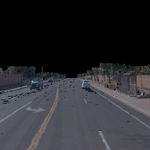}} & 
\T{\includegraphics[width=\imw]   {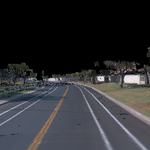}} & 
\T{\includegraphics[width=\imw]      {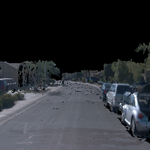}} & 
\T{\includegraphics[width=\imw]    {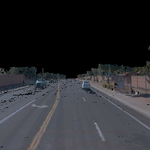}} & 
\T{\includegraphics[width=\imw]      {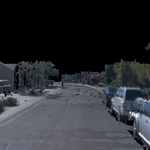}} & 
\T{\includegraphics[width=\imw]    {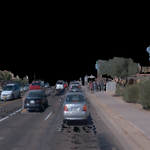}} \\
[+21pt]
\T{\includegraphics[width=\imw]     {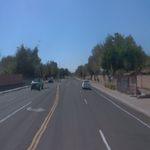}} & 
\T{\includegraphics[width=\imw]   {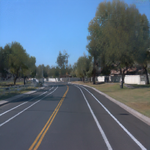}} & 
\T{\includegraphics[width=\imw]      {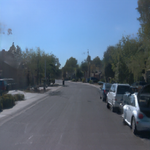}} & 
\T{\includegraphics[width=\imw]    {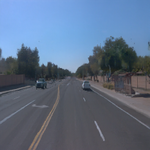}} & 
\T{\includegraphics[width=\imw]      {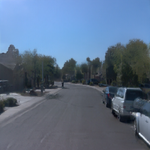}} & 
\T{\includegraphics[width=\imw]    {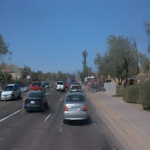}} \\
[+21pt]
&&&&\\

\T{\includegraphics[width=\imw]     {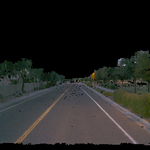}} & 
\T{\includegraphics[width=\imw]   {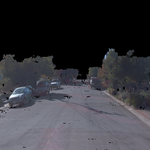}} & 
\T{\includegraphics[width=\imw]      {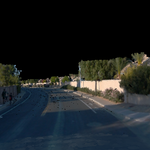}} & 
\T{\includegraphics[width=\imw]    {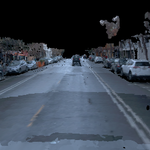}} & 
\T{\includegraphics[width=\imw]      {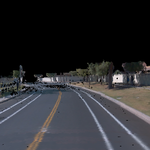}} & 
\T{\includegraphics[width=\imw]    {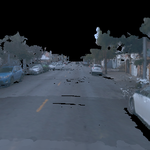}} \\
[+21pt]
\T{\includegraphics[width=\imw]     {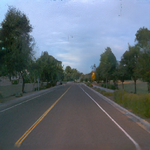}} & 
\T{\includegraphics[width=\imw]   {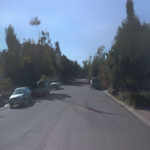}} & 
\T{\includegraphics[width=\imw]      {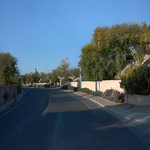}} & 
\T{\includegraphics[width=\imw]    {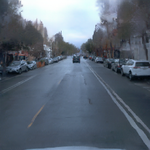}} & 
\T{\includegraphics[width=\imw]      {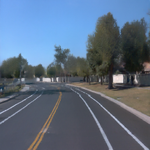}} & 
\T{\includegraphics[width=\imw]    {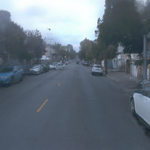}} \\
[+21pt]
&&&&\\

\T{\includegraphics[width=\imw]     {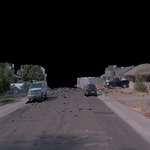}} & 
\T{\includegraphics[width=\imw]   {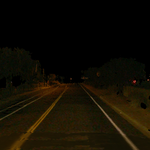}} & 
\T{\includegraphics[width=\imw]      {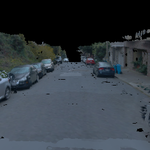}} & 
\T{\includegraphics[width=\imw]    {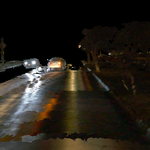}} & 
\T{\includegraphics[width=\imw]      {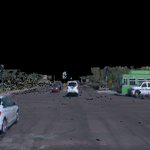}} & 
\T{\includegraphics[width=\imw]    {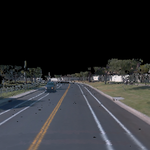}} \\
[+21pt]
\T{\includegraphics[width=\imw]     {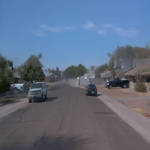}} & 
\T{\includegraphics[width=\imw]   {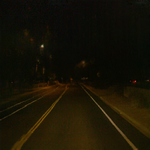}} & 
\T{\includegraphics[width=\imw]      {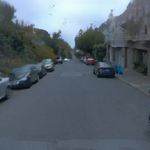}} & 
\T{\includegraphics[width=\imw]    {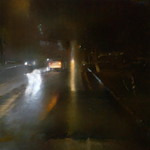}} & 
\T{\includegraphics[width=\imw]      {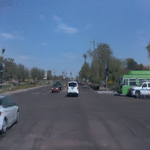}} & 
\T{\includegraphics[width=\imw]    {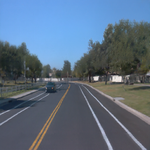}} \\
[+21pt]
&&&&\\

\end{tabular}
\caption{Qualitative results of SurfelGAN-SAC (1/2). We show pairs of surfel rendering and synthesized image.}
\label{fig:supp3}
\vspace{-0.15in}
\end{figure*}

\begin{figure*}
\centering
\footnotesize
\def\imh{0.15\textwidth}
\def\imw{0.15\textwidth}
\newcommand{\T}[1]{\raisebox{-0.5\height}{#1}}
\setlength{\tabcolsep}{1pt}
\begin{tabular}{cccccc}

\T{\includegraphics[width=\imw]     {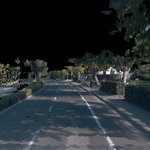}} & 
\T{\includegraphics[width=\imw]   {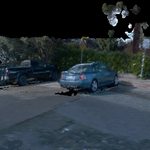}} & 
\T{\includegraphics[width=\imw]      {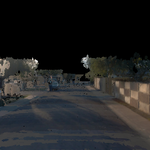}} & 
\T{\includegraphics[width=\imw]    {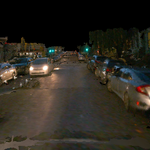}} & 
\T{\includegraphics[width=\imw]      {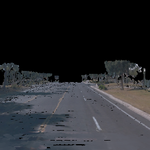}} & 
\T{\includegraphics[width=\imw]    {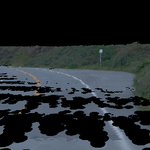}} \\
[+21pt]
\T{\includegraphics[width=\imw]     {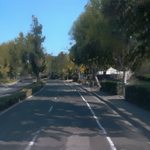}} & 
\T{\includegraphics[width=\imw]   {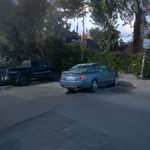}} & 
\T{\includegraphics[width=\imw]      {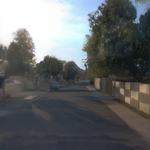}} & 
\T{\includegraphics[width=\imw]    {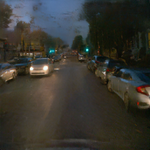}} & 
\T{\includegraphics[width=\imw]      {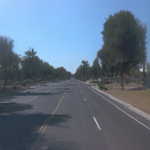}} & 
\T{\includegraphics[width=\imw]    {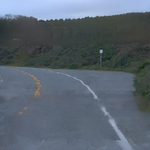}} \\
[+21pt]
&&&&\\

\T{\includegraphics[width=\imw]     {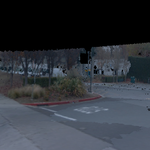}} & 
\T{\includegraphics[width=\imw]   {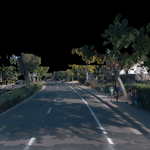}} & 
\T{\includegraphics[width=\imw]      {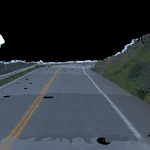}} & 
\T{\includegraphics[width=\imw]    {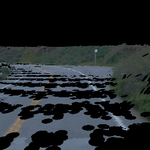}} & 
\T{\includegraphics[width=\imw]      {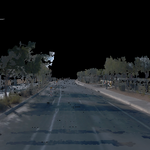}} & 
\T{\includegraphics[width=\imw]    {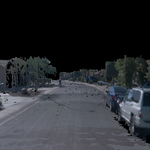}} \\
[+21pt]
\T{\includegraphics[width=\imw]     {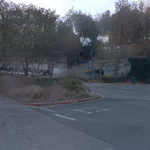}} & 
\T{\includegraphics[width=\imw]   {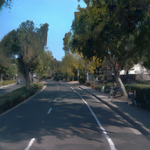}} & 
\T{\includegraphics[width=\imw]      {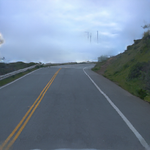}} & 
\T{\includegraphics[width=\imw]    {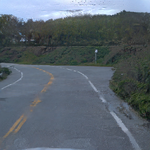}} & 
\T{\includegraphics[width=\imw]      {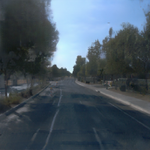}} & 
\T{\includegraphics[width=\imw]    {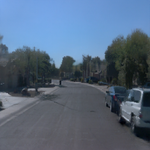}} \\
[+21pt]
&&&&\\

\T{\includegraphics[width=\imw]     {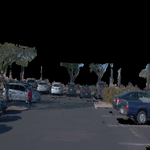}} & 
\T{\includegraphics[width=\imw]   {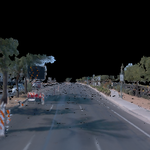}} & 
\T{\includegraphics[width=\imw]      {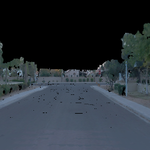}} & 
\T{\includegraphics[width=\imw]    {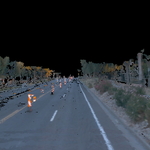}} & 
\T{\includegraphics[width=\imw]      {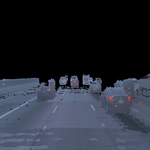}} & 
\T{\includegraphics[width=\imw]    {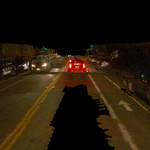}} \\
[+21pt]
\T{\includegraphics[width=\imw]     {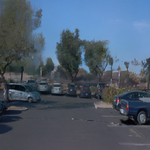}} & 
\T{\includegraphics[width=\imw]   {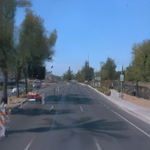}} & 
\T{\includegraphics[width=\imw]      {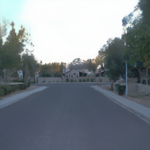}} & 
\T{\includegraphics[width=\imw]    {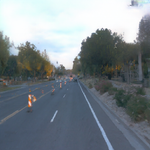}} & 
\T{\includegraphics[width=\imw]      {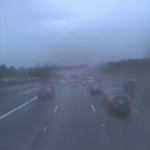}} & 
\T{\includegraphics[width=\imw]    {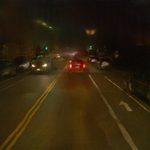}} \\
[+21pt]
&&&&\\

\T{\includegraphics[width=\imw]     {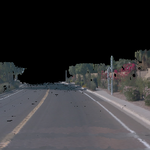}} & 
\T{\includegraphics[width=\imw]   {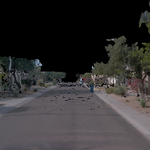}} & 
\T{\includegraphics[width=\imw]      {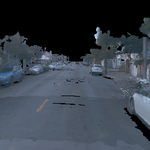}} & 
\T{\includegraphics[width=\imw]    {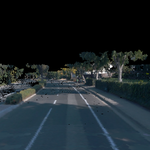}} & 
\T{\includegraphics[width=\imw]      {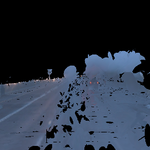}} & 
\T{\includegraphics[width=\imw]    {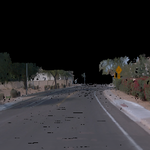}} \\
[+21pt]
\T{\includegraphics[width=\imw]     {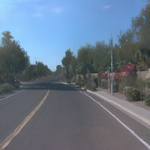}} & 
\T{\includegraphics[width=\imw]   {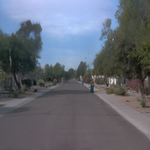}} & 
\T{\includegraphics[width=\imw]      {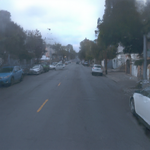}} & 
\T{\includegraphics[width=\imw]    {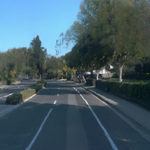}} & 
\T{\includegraphics[width=\imw]      {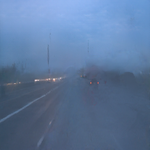}} & 
\T{\includegraphics[width=\imw]    {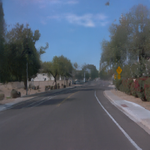}} \\
[+21pt]
&&&&\\

\end{tabular}
\caption{Qualitative results of SurfelGAN-SAC (2/2). We show surfel rendering and synthesized image pairs.}
\label{fig:supp3-2}
\vspace{-0.15in}
\end{figure*}

\begin{figure*}
\centering
\footnotesize
\def\imh{0.15\textwidth}
\def\imw{0.15\textwidth}
\newcommand{\T}[1]{\raisebox{-0.5\height}{#1}}
\setlength{\tabcolsep}{1pt}
\begin{tabular}{ccccc}

Surfel rendering & SurfelGAN-S & SurfelGAN-SA & SurfelGAN-SAC & REAL\\
\T{\includegraphics[width=\imw]     {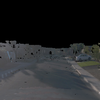}} & 
\T{\includegraphics[width=\imw]   {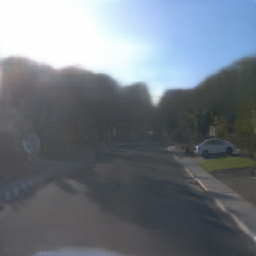}} & 
\T{\includegraphics[width=\imw]      {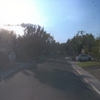}} & 
\T{\includegraphics[width=\imw]    {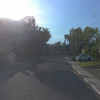}} & 
\T{\includegraphics[width=\imw]      {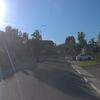}} \\
[+21pt]
&&&&\\

\T{\includegraphics[width=\imw]     {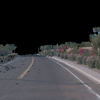}} & 
\T{\includegraphics[width=\imw]   {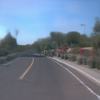}} & 
\T{\includegraphics[width=\imw]      {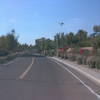}} & 
\T{\includegraphics[width=\imw]    {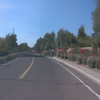}} & 
\T{\includegraphics[width=\imw]      {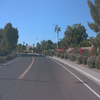}} \\
[+21pt]
&&&&\\

\T{\includegraphics[width=\imw]     {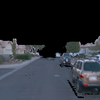}} & 
\T{\includegraphics[width=\imw]   {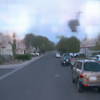}} & 
\T{\includegraphics[width=\imw]      {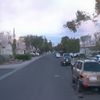}} & 
\T{\includegraphics[width=\imw]    {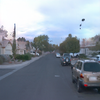}} & 
\T{\includegraphics[width=\imw]      {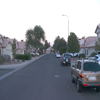}} \\
[+21pt]
&&&&\\

\T{\includegraphics[width=\imw]     {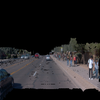}} & 
\T{\includegraphics[width=\imw]   {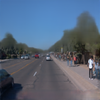}} & 
\T{\includegraphics[width=\imw]      {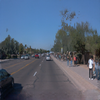}} & 
\T{\includegraphics[width=\imw]    {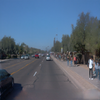}} & 
\T{\includegraphics[width=\imw]      {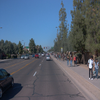}} \\
[+21pt]
&&&&\\

\T{\includegraphics[width=\imw]     {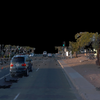}} & 
\T{\includegraphics[width=\imw]   {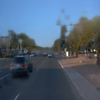}} & 
\T{\includegraphics[width=\imw]      {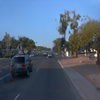}} & 
\T{\includegraphics[width=\imw]    {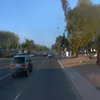}} & 
\T{\includegraphics[width=\imw]      {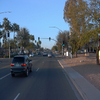}} \\
[+21pt]
&&&&\\

\T{\includegraphics[width=\imw]     {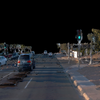}} & 
\T{\includegraphics[width=\imw]   {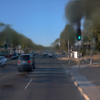}} & 
\T{\includegraphics[width=\imw]      {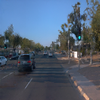}} & 
\T{\includegraphics[width=\imw]    {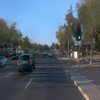}} & 
\T{\includegraphics[width=\imw]      {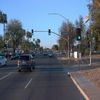}} \\
[+21pt]
&&&&\\

\T{\includegraphics[width=\imw]     {Figure/qualitative/012_surfel.png}} & 
\T{\includegraphics[width=\imw]   {Figure/qualitative/012_l1.png}} & 
\T{\includegraphics[width=\imw]      {Figure/qualitative/012_nocycle.png}} & 
\T{\includegraphics[width=\imw]    {Figure/qualitative/012_ours.png}} & 
\T{\includegraphics[width=\imw]      {Figure/qualitative/012_gt.png}} \\
[+21pt]
&&&&\\

\end{tabular}
\caption{Qualitative comparison between different SurfelGAN variants on WOD-EVAL.}
\label{fig:supp1}
\vspace{-0.15in}
\end{figure*}

\begin{figure*}
\centering
\footnotesize
\def\imh{0.15\textwidth}
\def\imw{0.15\textwidth}
\newcommand{\T}[1]{\raisebox{-0.5\height}{#1}}
\setlength{\tabcolsep}{1pt}
\begin{tabular}{cccccc}

\T{\includegraphics[width=\imw]     {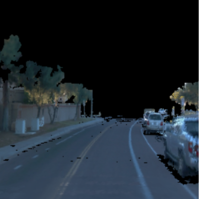}} & 
\T{\includegraphics[width=\imw]   {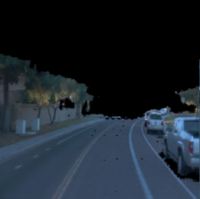}} & 
\T{\includegraphics[width=\imw]      
 {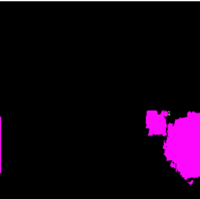}} & 
\T{\includegraphics[width=\imw]    {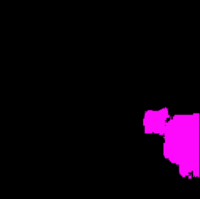}} &
\T{\includegraphics[width=\imw]   
{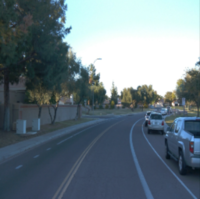}} & 
\T{\includegraphics[width=\imw]    {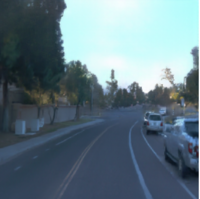}} \\
[+21pt]
&&&&\\

\T{\includegraphics[width=\imw]     {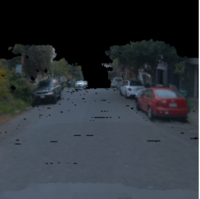}} & 
\T{\includegraphics[width=\imw]   {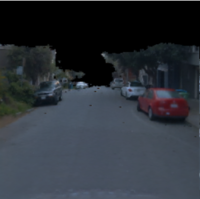}} & 
\T{\includegraphics[width=\imw]      
 {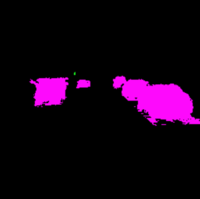}} & 
\T{\includegraphics[width=\imw]    {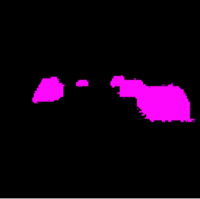}} &
\T{\includegraphics[width=\imw]     
{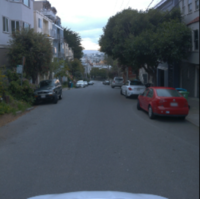}} & 
\T{\includegraphics[width=\imw]    {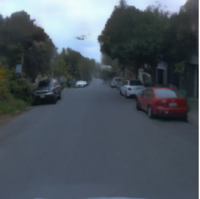}} \\
[+21pt]
&&&&\\

\T{\includegraphics[width=\imw]     {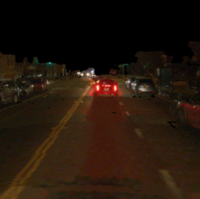}} & 
\T{\includegraphics[width=\imw]   {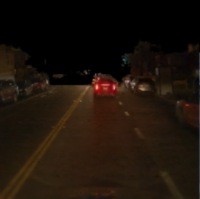}} & 
\T{\includegraphics[width=\imw]      {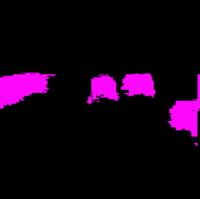}} & 
\T{\includegraphics[width=\imw]    {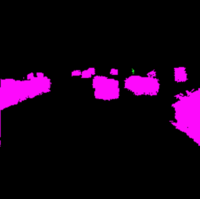}} & 
\T{\includegraphics[width=\imw]      {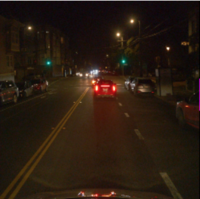}} & 
\T{\includegraphics[width=\imw]    {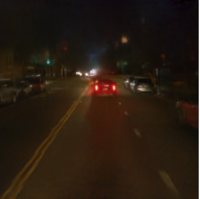}} \\
[+21pt]
&&&&\\

\T{\includegraphics[width=\imw]     {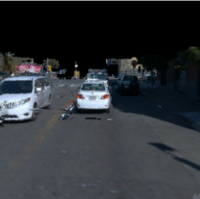}} & 
\T{\includegraphics[width=\imw]   {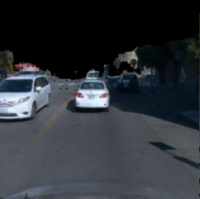}} & 

\T{\includegraphics[width=\imw]      {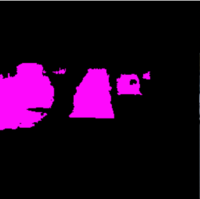}} & 
\T{\includegraphics[width=\imw]    {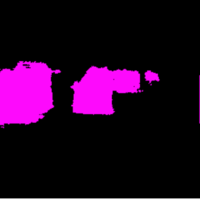}} & 
\T{\includegraphics[width=\imw]      {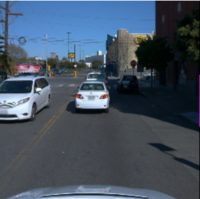}} & 
\T{\includegraphics[width=\imw]    {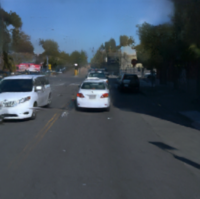}} \\
[+21pt]
&&&&\\

\T{\includegraphics[width=\imw]     {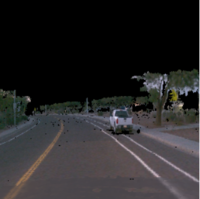}} & 
\T{\includegraphics[width=\imw]   {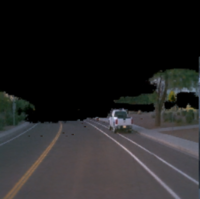}} & 
\T{\includegraphics[width=\imw]      {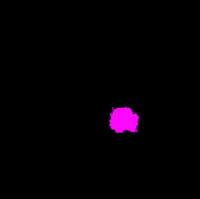}} & 
\T{\includegraphics[width=\imw]    {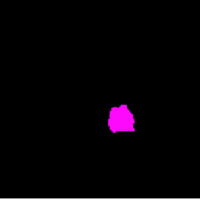}} & 
\T{\includegraphics[width=\imw]      {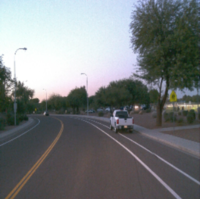}} & 
\T{\includegraphics[width=\imw]    {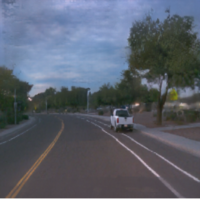}} \\
[+21pt]
&&&&\\

\T{\includegraphics[width=\imw]     {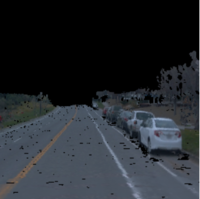}} & 
\T{\includegraphics[width=\imw]   {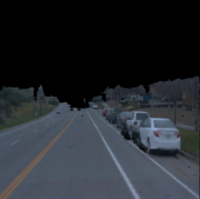}} & 
\T{\includegraphics[width=\imw]      {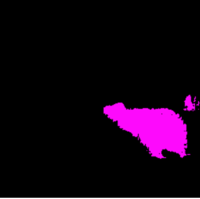}} & 
\T{\includegraphics[width=\imw]    {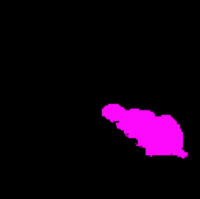}} & 
\T{\includegraphics[width=\imw]      {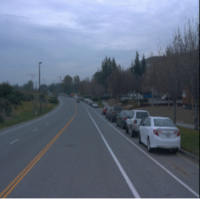}} & 
\T{\includegraphics[width=\imw]    {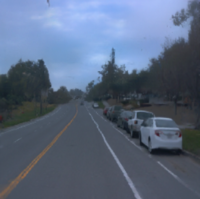}} \\
[+21pt]
&&&&\\

\end{tabular}
\caption{Visualization of the different outputs of SurfelGAN-SAC. From left to right: surfel rendering, generated surfel rendering, semantic map, generated semantic map, camera image, generated camera image.}
\label{fig:supp2}
\vspace{-0.15in}
\end{figure*}

We provide additional qualitative results:~\figref{fig:supp4} contains synthesized output when gradually perturbing the current viewpoint. \figref{fig:supp5} contains synthesized output when we also perturb the other objects in the scene. \figref{fig:supp1} provides a visual comparison between different model variants. Results of generated surfel images and semantic maps of SurfelGAN-SAC can be found in~\figref{fig:supp2}. More visual results of SurfelGAN-SAC can be find in~\figref{fig:supp3} and~\figref{fig:supp3-2}.

\section{Analysis on Surfel Image Coverage}
We examine the object detection metric for different surfel image coverage ratios. Specifically, we define the surfel image coverage ratio as the percentage of area in the surfel rendered image that is non-empty. The results can be found in~\tblref{tab:breakdown:ratio}. We found that our SurfelGAN model performs similarly when the coverage ratio is above $30\%$, but the performance declines significantly when the input surfel map becomes more and more incomplete. This can happen when the SDV is looking in a direction where the surfel scene reconstruction is incomplete, for example, if it is positioned at the surfel map boundary looking outward. This observation suggests that building more complete surfel scene maps over multiple runs should be helpful.

\begin{table}[!tbp]
  \footnotesize
  \centering
  \begin{tabular}{lcccc}
  \toprule
  Perturbation & AP@50 & AP@75 & AP\\
  \hline
  $r \leq 0.3$ & 0.490 & 0.155 & 0.218\\
  $0.3 < r \leq 0.5$  & 0.577 & 0.235 & 0.279\\
  $0.5 < r$  & 0.566 & 0.172 & 0.253\\
  \bottomrule
  \end{tabular}
 \caption{Object detection metric for different surfel image coverage ratios (r) on WOD-EVAL-NV.}
 \label{tab:breakdown:ratio}
\end{table}

\end{document}